\documentclass[a4paper,conference]{IEEEtran}
\usepackage{subfigure}
\usepackage{graphicx}
\usepackage{CJKutf8}  
\usepackage{pinyin}
\usepackage{enumerate}
\usepackage{bm}
\usepackage{multirow}

\usepackage{soul}
\usepackage{url}
\usepackage[hidelinks]{hyperref}
\usepackage[utf8]{inputenc}
\usepackage[small]{caption}
\usepackage{graphicx}
\usepackage{amsmath}
\usepackage{amsthm}
\usepackage{booktabs}
\usepackage{algorithm}
\usepackage{algorithmic}
\usepackage{color}
\usepackage{cite}
\ifCLASSINFOpdf
\else
\fi

\begin{document}
\IEEEoverridecommandlockouts
\begin{CJK}{UTF8}{gbsn}

\title{Moto: Enhancing Embedding with Multiple Joint Factors for Chinese Text Classification}






\author{\IEEEauthorblockN{Xunzhu Tang\IEEEauthorrefmark{1},
Rujie Zhu\IEEEauthorrefmark{2},
Tiezhu Sun\IEEEauthorrefmark{3} and
Shi Wang\IEEEauthorrefmark{4}}
\IEEEauthorblockA{\IEEEauthorrefmark{1} Huazhong University \\of Science and Technology\\Wuhan, China\\ Email: realdanieltang@gmail.com}
\IEEEauthorblockA{\IEEEauthorrefmark{3}University of Central Florida, Orlando, FL, USA\\ Email: rujie.zhu@ucf.edu}
\IEEEauthorblockA{\IEEEauthorrefmark{2} Momenta, Suzhou, China\\
Email: tiezhu.sun@uni.lu}
\IEEEauthorblockA{\IEEEauthorrefmark{4}Institute of Computing Technology, Chinese Academy, Beijing, China \\Email:wangshi@ict.ac.cn}}

\maketitle

\begin{abstract}
Recently, language representation techniques have achieved great performances in text classification. However, most existing representation models are specifically designed for English materials, which may fail in Chinese because of the huge difference between these two languages. Actually, few existing methods for Chinese text classification process texts at a single level. However, as a special kind of hieroglyphics, radicals of Chinese characters are good semantic carriers. In addition, Pinyin codes carry the semantic of tones, and Wubi reflects the stroke structure information, \textit{etc}. Unfortunately, previous researches neglected to find an effective way to distill the useful parts of these four factors and to fuse them. In our works, we propose a novel model called Moto: Enhancing Embedding with \textbf{M}ultiple J\textbf{o}int Fac\textbf{to}rs. Specifically, we design an attention mechanism to distill the useful parts by fusing the four-level information above more effectively. We conduct extensive experiments on four popular tasks. The empirical results show that our Moto achieves SOTA 0.8316 ($F_1$-score, 2.11\% improvement) on Chinese news titles, 96.38 (1.24\% improvement) on Fudan Corpus and 0.9633 (3.26\% improvement) on THUCNews.
\end{abstract}


\section{Introduction}
Different from the English-like languages whose words can be spelt out according to their pronunciation and meanings are related with words themselves, semantics of Chinese are relevant to characters and highly associated with component parts ({\it i.e.,} radicals \cite{shi2015radical,tao2019radical}), structure of characters ({\it i.e.,} Wubi codes \cite{zhou2019multiple}), and tones of pronunciation ({\it i.e.,} Pinyin codes \cite{chen2015neural}). Over the past years, many works applied only one of different aspects of Chinese characters on sequence-to-sequence model to enhance the ability of capturing semantic features. We advocate fusing all the four aspects of Chinese characters ({\it i.e.,} multiple joint factors) to enhance Chinese Embedding, which would bring much better performance for Chinese texts classification \cite{peng2003text}. 

\begin{figure*}[!ht]
\centering
\subfigure[Radical]{
\centering
\includegraphics[width=5.15cm]{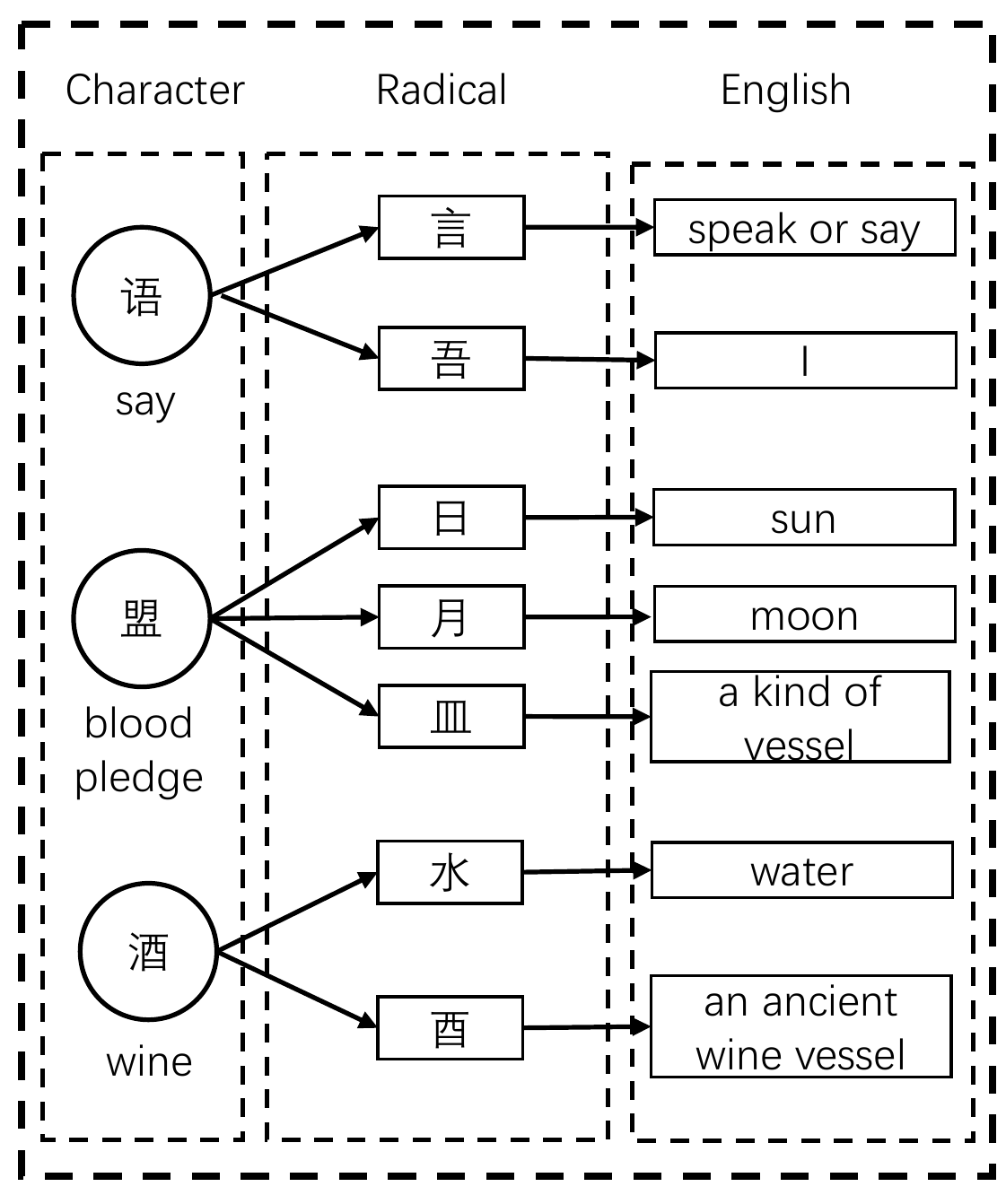}
\label{lb:radical}
}%
\subfigure[Wubi]{
\centering
\includegraphics[width=5.85cm]{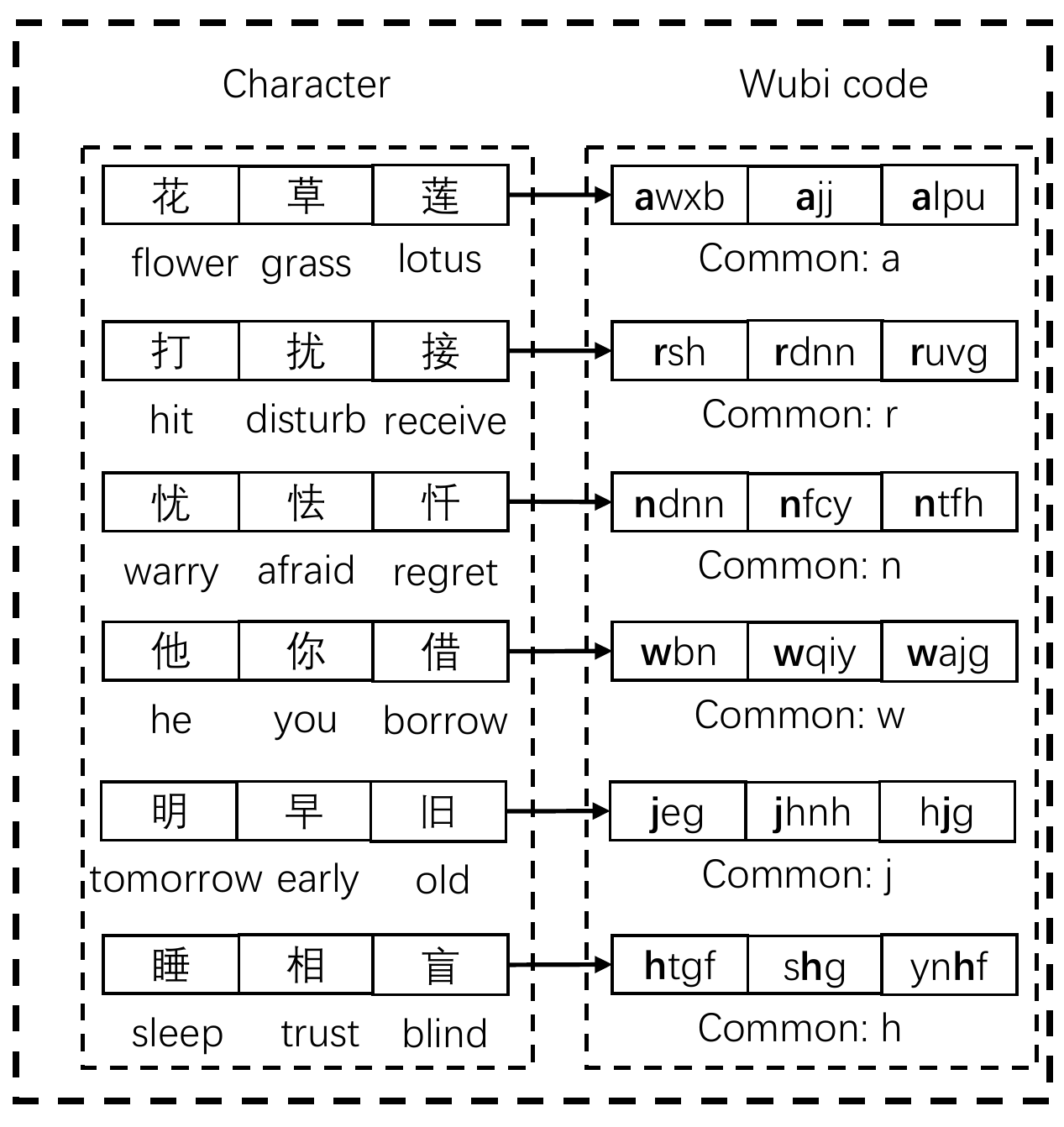}
\label{lb:wubi}
}%
\subfigure[Pinyin]{
\centering
\includegraphics[width=4cm]{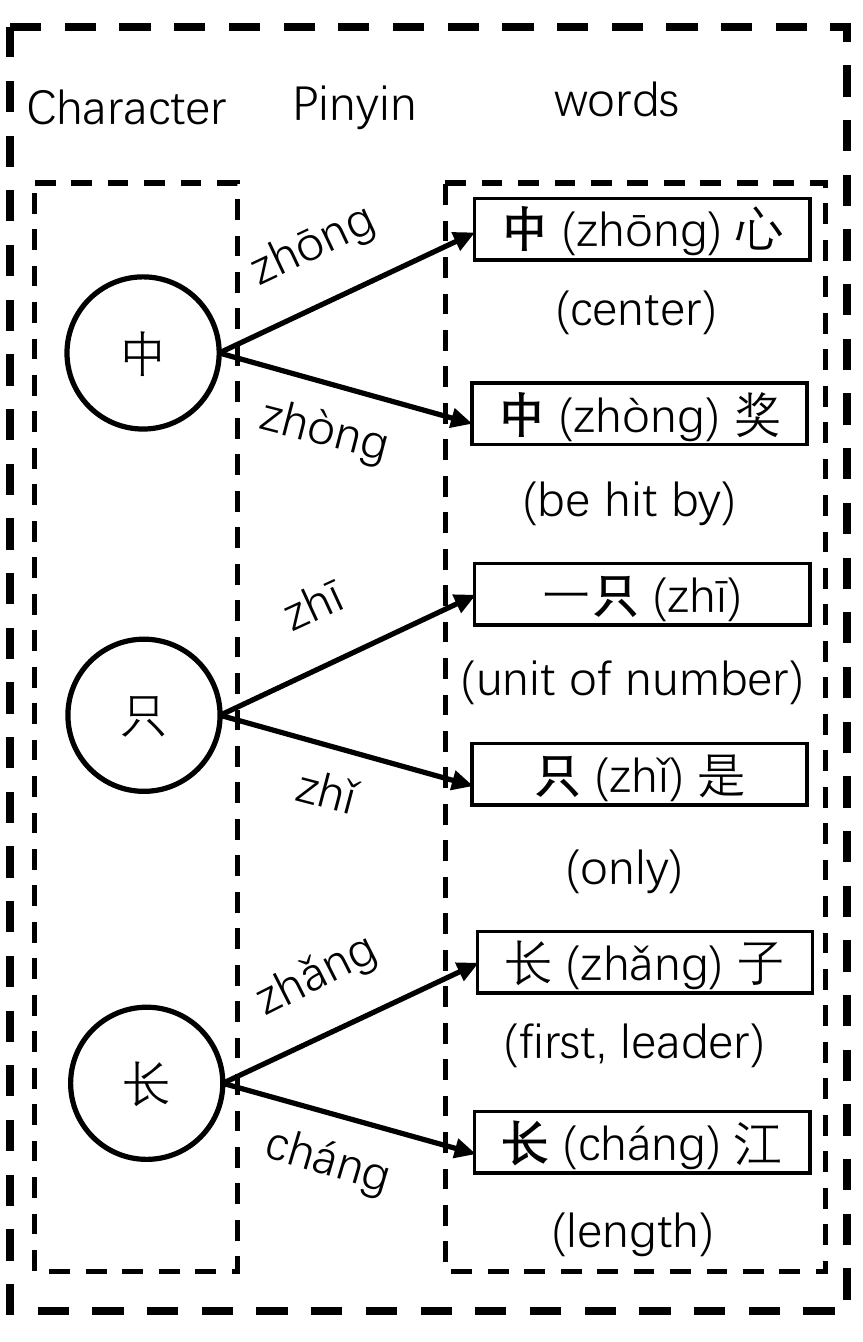}
\label{lb:pinyin}
}%
\centering
\caption{We employ three kind of representations to enhance the character embedding. Figure (a) indicates that radicals can show more details of characters; Figure (b) shows that Wubi code can capture the structure information of characters; And figure (c) expresses that the Pinyin (with pronounce) is important to Chinese characters.}
\label{fig:cp}
\end{figure*}

As a kind of pictograph language, the uniqueness of Chinese is that its character system is based on hieroglyphics, which means that Chinese characters have their raw meanings. In other words, not only Chinese characters themselves can express specific meanings, but also their component parts are important carriers of semantics, which is the main point where Chinese differ from English. As shown in Figure \ref{lb:radical}, `盟' (blood pledge) consists three radicals : `日 (sun)', `月' (moon), and `皿' (a kind of vessel). Furthermore, the positions of these radicals in characters are also significant in hieroglyphics. For example, as shown in Figure \ref{lb:wubi}, `花' (flower), `草' (grass), and `莲' (lotus) have one common radical `艹', the same structure (upper-down), and the same position ({\it i.e.,} `艹' is in the upper position, which was recorded as '{\it a}' in Wubi codes), which means they are all plants. It is not difficult to see that radicals and Wubi codes could help us to recognize semantics for classifying Chinese texts. In addition, Pinyin codes could also help us to capture the semantic of tone, as shown in Figure \ref{lb:pinyin}.

Additionally, there has been a lot of works aiming to employ Wubi codes on Chinese Word Segmentation (CWS) \cite{zhou2019multiple}, radicals on Chinese text classification \cite{shi2015radical,tao2019radical}, and Pinyin codes on Chinese embeddings \cite{chen2015neural}. However, these works simply use single aspect of characters like radicals to enhance character-level embeddings. 

Inspired by the importance of radicals of characters, Wubi codes, and Pinyin codes, we conduct an explorative study in Chinese text classification with attention mechanism to jointly leverage four granularities of features, which we call Moto in this paper. The main contributions are threefold:

\begin{itemize}
    \item We first employ the attention mechanism to capture the effective parts four-granularity features ({\it i.e., }characters, radicals, Wubi codes, and Pinyin codes.) 
    \item We first design a novel mechanism to confirm the weights among these four-granularity features dynamically.
    \item We conduct extensive experiments on four real-world and public datasets in four granularities respectively, and demonstrate the effectiveness of characters, radicals, Wubi codes, and Pinyin codes. 
\end{itemize}

\section{Multiple Embeddings}
In this section, we will discuss the priority of employing four different granularities, including characters, radicals, Wubi codes, and Pinyin codes, as shown in Figure \ref{fig:architecture}. 

\textbf{Character} is usually recognized as the smallest unit to process Chinese text classification. Recent works show that character embeddings are the most fundamental inputs for neural networks \cite{chen2015long,cai2016neural,cai2017fast,zhou2019multiple} since it is easy to learn contextual information with sequence-to-sequence models. However, Chinese character system is based on hieroglyphics, whose component parts of characters are also the carriers of semantics. 

\textbf{Radical} or radical-like components serving as the basic units for building Chinese characters has been explored in \cite{li2015component,shi2015radical}. Commonly, radicals has the following two features. The first one is that one radical normally has one or two types. `言' (speak) is itself, but it becomes `讠' in character `语'. The second is that radicals have specific meanings. In Figure \ref{lb:radical}, the character `语' ({\it talk or speak}) have radicals : `讠' ({\it the same as 言, speak}) and `吾' ({\it I}). Obviously, radical provides extra photographic features of characters. 

\textbf{Wubi} is another effective representation of Chinese characters, which includes more comprehensive structure information compared to radical. Each element in a Wubi code represents a type of structure (or stroke) in characters. In Figure \ref{lb:wubi}, `花' (flower), `草' (grass), and `莲' (lotus) are all related to plants, and their Wubi codes `awxb', `ajj', and `alpu' have one common letter `{\it a}', which is corresponding to radical `艹'. Therefore, Wubi is an efficient approach to capture structure features of Chinese characters.

\textbf{Pinyin} is a English-like expression approach of Chinese characters. Besides, Pinyin is highly relevant to semantics - one character may have multiple pronunciations corresponding to different semantic meanings \cite{zhou2019multiple}, which is called polyphone in Chinese. Figure \ref{lb:pinyin} shows several polyphone characters. `中' has two pronunciations (Pinyin). When pronounced as `zhōng' in `中心', it means `center'. However, when it refer to the meaning of `be hit by' when pronounced in `中奖'. Obviously, it is beneficial for Chinese characters to use Pinyin code to capture phonetic information. 

\section{Details of Moto Model} \label{sec:RL}
\begin{figure*}[!ht]
    \centering
    \includegraphics[scale=0.5]{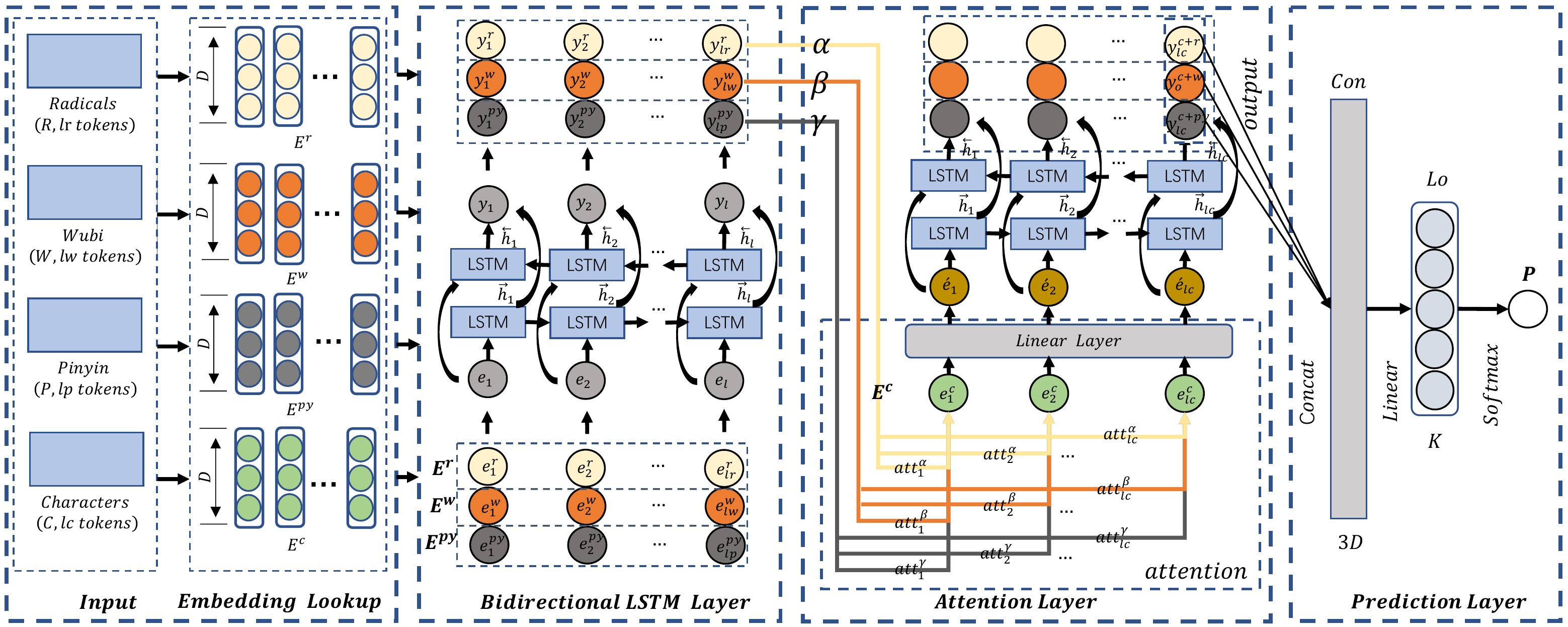}
    \caption{Network architecture of \textbf{Moto}, including four-granularity representations of Chinese: Radicals, Wubi, Pinyin, and Characters.}
    \label{fig:architecture}
\end{figure*}
In this section, we introduce our joint enhanced character embedding model (Moto), which utilizes radical, Wubi code, and Pinyin as supplementary input text. The challenge is that how to distill the important parts of these factors and how to conform the weights among them. The paper extends the method of attention mechanism \cite{vaswani2017attention} to infer the weights, which will be discussed in \ref{sec:AL}. As shown in Figure \ref{fig:architecture}, Moto mainly contains four parts: {\it Input Layer}, {\it Bidirectional LSTM Layer}, {\it Attention Layer}, and {\it Prediction Layer}.
\vspace{-0.3cm}
\subsection{Input Layer} \label{sec: IL}
Given a Chinese-character sequence \textit{C} which contains \textit{lc} characters, \emph{i.e.}, {\it C =} \{$c_1, c_2,\dots, c_{lc}$\}, where each character $c_i$ (1 $\leq$ {\it i} $\leq$ {\it lc (length of characters)}) is an independent item in {\it C}. Meanwhile, {\it C} will be mapped into radicals, Wubi, and Pinyin respectively by the usage of {\it Open Chinese dictionary } \footnote{http://www.kaifangcidian.com/han/chaizi}, Wubi Library \footnote{https://github.com/sfyc23/python-wubi}, and Pypinyin Library \footnote{https://pypi.org/project/pypinyin/}, \emph{i.e.}, \textit{lr} (the length of radicals) radical-level radicals {\it R=}\{$r_1, r_2, \dots, r_{lr} $\}, \textit{lw} (the length of Wubi codes) Wubi codes {\it W=}\{$w_1, w_2, \dots, w_{lw} $\}, and \textit{lp} (the length of Pinyin codes) Pinyin codes {\it Py=}\{$py_1, py_2, \dots, py_{lp} $\}. Then we retrieve four granularities of features (\emph{i.e., }{\it C, R, W, Py}) and obtain four embedding matrices using word2vec tool \footnote{https://radimrehurek.com/gensim/} \cite{mikolov2013distributed}. As shown in Figure \ref{fig:architecture}, the embedding of sequence {\it C} can be represented of $E^c$= \{$e^c_1, e^c_2, \dots, e^c_{lc},$\} (where $e^c_i$ is the representation of $c_i$). Similarly, $E^r$= \{$e^r_1, e^r_2, \dots, e^r_{lr},$\} (where $e^r_i$ is the representation of $r_i$), $E^w$= \{$e^w_1, e^w_2, \dots, e^w_{lw},$\} (where $e^c_i$ is the representation of $c_i$), $E^{py}$= \{$e^{py}_1, e^{py}_2, \dots, e^{py}_{lp},$\} (where $e^{py}_i$ is the representation of $py_i$). For simplicity, we set the vector dimension of each level embeddings as {\it D}, which means $E^c \in R^{lc \times D}$, $E^r \in R^{lr \times D}$, $E^w \in R^{lw \times D}$, $E^{py} \in R^{lp \times D}$. Then we feed them in BiLSTM layer directly.

\subsection{Bidirectional LSTM Layer} \label{sec:BLL}
In this section, we employ BiLSTM to capture contextual information of input sequence and obtain independent-contextual representation. LSTM \cite{hochreiter1997long} is an advanced version of recurrent neural network (RNN) with extra forget and memory which are employed to alleviate the gradient vanishing problem and keep the term information as long as possible. Given a specific feature embedding sequence {\it E=} \{$e_1, e_2, \dots, e_n$\}, the whole progress in the forward LSTM is calculated as follows:

\begin{small}
\begin{equation}
\left[ \begin{array}{ccc}
\overrightarrow{i_t}\\
\overrightarrow{f_t}\\
\overrightarrow{o_t}\\
\overrightarrow{\widetilde{c_t}}
\end{array}
\right ]=
\left[ \begin{array}{ccc}
\sigma\\
\sigma\\
\sigma\\
\tanh
\end{array} 
\right ]
\Bigg(
W^T
\left[ \begin{array}{ccc}
e_t\\
\overrightarrow{h_{t-1}}
\end{array}
\right ]
+b
\Bigg)
\end{equation}
\end{small}

\vspace{-0.3cm}

\begin{small}
\begin{equation}
    \begin{split}
        \overrightarrow{c_t}&=f_t\ast\overrightarrow{c_{t-1}}+\overrightarrow{i_t}\ast\overrightarrow{\widetilde{c_t}} \\
        \overrightarrow{h_t}&=\overrightarrow{o_t}\ast\tanh{\overrightarrow{c_t}}
    \end{split}
\end{equation}
\end{small}
where $i_t, f_t,$ $o_t,$ and $\widetilde{c_t}$ denote a set of input, forget, output, and new layer to update current information $c_t$ respectively. Moreover, $W$ equals to the concatenation of $W_i$ (a matrix parameter in input gate), $W_f$ (a matrix parameter in forget gate), $W_o$ (a matrix parameter in output gate), and $W_c$ (a matrix parameter in {\it tanh} layer). The progress above can be described as equation $W$ = $W_i$ $\oplus$ $W_f$ $\oplus$ $W_o$ $\oplus$ $W_c$, where symbol $\oplus$ represents the concatenation function. Similar to $W$, $b$= $b_i$ $\oplus$ $b_f$ $\oplus$ $b_o$ $\oplus$ $b_c$. In addition, symbol $\sigma(\cdot)$ indicates the sigmoid function. Similar to forward LSTM, the hidden state of {\it t}-th step in the backward LSTM can be represented as $\overleftarrow{h_t}$. Then we concatenate $\overrightarrow{h_t}$ and $\overleftarrow{h_t}$ as $y_t$, which is the hidden output of each BiLSTM cell at the \textit{t}-th step, and the process is computed by $y_t$ = $\overrightarrow{h_t}$$\oplus$ $\overleftarrow{h_t}$.

As shown in Figure \ref{fig:architecture}, there are three input embeddings in the Bidirectional LSTM Layer (\emph{i.e.}, $E^r$, $E^w$, and $E^{py}$), which will be fed to three different BiLSTM network (\emph{i.e.}, $BiLSTM^r$, $BiLSTM^w$, and $BiLSTM^{py}$ which share parameters). Then we can get three related outputs from these BiLSTM networks, \emph{i.e.,} $Y^r$=\{$y^r_1, y^r_2, \dots, y^r_{lr}$\}, $Y^w$=\{$y^w_1, y^w_2, \dots, y^w_{lr}$\} and $Y^{py}$=\{$y^{py}_1, y^{py}_2, \dots, y^{py}_{lr}$\}, which will be taken into calculation in Section \ref{sec:AL}.

\subsection{Attention Layer} \label{sec:AL}
BiLSTMs provide three outputs of $Y^r$, $Y^w$, and $Y^{py}$ in the last section. For a specific output $Y^r$, different element $y^r_i$ $\in$ $Y^r$ (where 1 $\leq$ {\it i} $\leq$ {\it lr}) has different effect on Chinese character-level semantics. In this section, we design an attention mechanism to calculate the weight of different element $y^r_i$ in $Y^r$ at the $\textit{j}$-th ${BiLSTM}^c$ cell. The $\bm{Attention Layer}$ in Figure \ref{fig:architecture} shows that each time of character-level BiLSTM network has one attention input, \emph{i.e.,} $att^{\alpha}_i$ $\in$ $att^{\alpha}$= \{$att^{\alpha}_1$, $att^{\alpha}_2$, \dots, $att^{\alpha}_{lc}$\}, $att^{\beta}_i$ $\in$ $att^{\beta}$= \{$att^{\beta}_1$, $att^{\beta}_2$, \dots, $att^{\beta}_{lc}$\}, and $att^{\gamma}_i$ $\in$ $att^{\gamma}$= \{$att^{\gamma}_1$, $att^{\gamma}_2$, \dots,  $att^{\gamma}_{lc}$\}. 

For ease of exposition, we take $att^{\alpha}$ as the main example in the next phase of inference. The sequence network to deal with character-level embeddings $\bm{E^c}$ is donated as {\it $BiLSTM^c$}. In each time $BiLSTM^c$ receives a vector of embedding of $y^c_j$ $\in$ $Y^c$= \{$y^c_1$, $y^c_2$, \dots, $y^c_{lc}$\}, where $y^c_j$= $\overrightarrow{h^c_j}$ $\oplus$ $\overleftarrow{h^c_j}$. 

In order to get weight of element $y^r_i$ in $Y^r$ at \textit{j}-th time of $BiLSTM^c$, we utilize dot function to figure out relevance between $y^r_i$ and $y^c_{j+1}$. We denote the relevance between $y^r_i$ and $y^c_j$ as $re_{i,j}$, and the process of calculation is as follows:

\begin{small}
\begin{equation}
    \begin{split}
        re_{i,j}&= {y^r_i}^Ty^c_j \\
        re_{:, j}&= \{re_{1, j}, re_{2, j}, \dots, re_{lr, j}\}
    \end{split}
\end{equation}
\end{small}
Then we employ {\it Softmax} to normalize $\bm{re_{:,j}}$ to get the attention distribution $\alpha$=\{$\alpha_{1, j}$, $\alpha_{2, j}$, \dots, $\alpha_{lr, j}$\}, where $\alpha_{i, j}$ is calculated as Equation \ref{eq:alpha}.

\begin{small}
\begin{equation} \label{eq:alpha}
    \begin{split}
        \alpha_{i, j}&= \frac{\exp{(re_{i, j})}}{\sum_{k=1}^{lr}\exp{(re_{k, j})}}, where \sum_{i=1}^{lr}\alpha_i = 1 \\
    \end{split}
\end{equation}
\end{small}

As a result, all weights of \{$y^r_1$, $y^r_1$, \dots, $y^r_{lr}$\} have been figured out. Here we use weighted arithmetic to value the whole effect of radical-level output $Y^r$ on character-level embedding. The calculation of weighted value and concatenation are as follows:

\begin{small}
\begin{equation} \label{eq:mixalpha}
    \begin{split}
        att^{\alpha}_j &= \sum_{i=1}^{lr}\alpha_{i, j}\times y_i^r \\
        \Acute{e}^c_{j+1} &= Linear(att^{\alpha}_j \oplus e^c_{j+1})
    \end{split}
\end{equation}
\end{small}
where $att^{\alpha}_j$ is the weighted effect of radical-level embedding in \textit{j}-th output of ${BiLSTM}^c$ cell. 

After the attention operation, the corresponding attention of radical-level embeddings to each origin input $e^c_j$ is available, \emph{i.e.,} $att^{\alpha}$= \{$att^{\alpha}_1$, $att^{\alpha}_2$, \dots, $att^{\alpha}_{lc}$\}. Instead of input $e^c_{j+1}$, we feed the new input $\Acute{e}^c_{j+1}$ into the neural network, where $\Acute{e}^c_{j+1}$ $\in$ $\Acute{E}^c$= \{$\Acute{e}^c_1$, $\Acute{e}^c_2$, \dots, $\Acute{e}^c_{lc}$\}. After decades of epochs of training ${BiLSTM}^c$ network with input $\Acute{E}^c$, it outputs a contextual sequence $Y^{c+r}$= \{$y^{c+r}_1$, $y^{c+r}_2$, \dots, $y^{c+r}_{lc}$\}. 

Similar to the attention operation of radical-level embedding on characters, Moto figures out that the output of fusion information of Wubi and Pinyin on characters respectively, $\emph{i.e.,}$ $Y^{c+w}$= \{$y^{c+w}_1$, $y^{c+r}_2$, \dots , $y^{c+w}_{lc}$\} and $Y^{c+py}$= \{$y^{c+py}_1$, $y^{c+py}_2$, \dots, $y^{c+py}_{lc}$\}. 
\subsection{Prediction Layer} \label{sec:PL}
Last but not least, we employ the last items of $Y^{c+py}$, $Y^{c+py}$ and $Y^{c+py}$ ( $\emph{i.e.,}$ $y^{c+py}_{lc}$, $y^{c+py}_{lc}$ and $y^{c+py}_{lc}$) as the final output, then we conduct a concatenation operation on them and get a comprehensive representation {\it Con} $\in$ $R^{3D}$ ({\it i.e.,} \textit{Con}= $y^{c+r+w+py}_{lc}$ = $y^{c+r}_{lc}$ $\oplus$ $y^{c+w}_{lc}$ $\oplus$ $y^{c+py}_{lc}$). After that, we feed {\it Con} into a fully-connected neural network to obtain an output vector {\it Lo} $\in$ $R^K$ (${\it Lo}$= \{$Lo_1$, $Lo_2$, \dots, $Lo_{K}$\}, and {\it K} is the number of classes in classification task), the whole calculation is shown as follows:

\begin{small}
\begin{equation} 
    \begin{split}
        Lo &= \sigma(Con \times W)\\
    \end{split}
\end{equation}
\end{small}
where {\it W} $\in$ {\it $R^{3D\times K}$} is the weight matrix for dimension transformation, and $\sigma(\cdot)$ is an activation function named {\it sigmoid}. 

Finally, we utilize a {\it Softmax} layer to map each element in {\it Lo} to a conditional probability. The calculation of probability distribution and class index corresponding to the max one are shown as follows:

\begin{equation} 
    \begin{split}
        p^{Lo_i}&= \frac{\exp{(Lo_i)}}{\sum_{k=1}^K\exp{(Lo_i)}}\\
        P &= \arg\max(P^{Lo}) 
    \end{split}
\end{equation}
where $\sum_{i=1}^K p^{Lo_i}$= 1 and $P^{Lo}$= \{$p^{Lo_1}, p^{Lo_1},$ \dots, $p^{Lo_K}$\}.

\begin{table*}[!ht]
\caption{Experimental results of different methods on Chinese news titles, Fudan Corpus, Douban movie review, and THUCNews.}
\centering
\begin{tabular}{|c|c|c|c|c|}
\hline
\multirow{2}{*}{Methods} & \begin{tabular}[c]{@{}c@{}}Chinese news titles\\ dataset \#1\end{tabular} & \begin{tabular}[c]{@{}c@{}}Chinese news titles\\ dataset \#2\end{tabular} & Fudan Corpus & THUCNews \\ \cline{2-5} 
 & F1(P,R) & F1(P,R) & F1(P,R) & F1(P,R) \\ \hline
\begin{tabular}[c]{@{}c@{}}SVM+BOW(C)\\ SVM+BOW(R)\\ SVM+BOW(W)\\ SVM+BOW(Py)\end{tabular} & \begin{tabular}[c]{@{}c@{}}0.7421 (0.7440, 0.7420)\\ 0.4697 (0.4652, 0.4809)\\ 0.6021 (0.6041, 0.6002)\\ 0.7290 (0.7309, 0.7271)\end{tabular} & \begin{tabular}[c]{@{}c@{}}0.7252 (0.7268, 0.7255)\\ 0.4691 (0.4636, 0.4813)\\ 0.4852 (0.4783, 0.4923)\\ 0.6702 (0.6874, 0.6539)\end{tabular} & \begin{tabular}[c]{@{}c@{}}0.8434 (0.8373, 0.8495)\\ 0.8187 (0.8216, 0.8158)\\ 0.8303 (0.8229, 0.8378)\\ 0.8359 (0.8367, 0.8352)\end{tabular} & \begin{tabular}[c]{@{}c@{}}0.8713 (0.8811, 0.8618)\\ 0.8641 (0.8637, 0.8646)\\ 0.8638 (0.8597, 0.8679)\\ 0.8703 (0.8778, 0.8629)\end{tabular} \\ \hline
\begin{tabular}[c]{@{}c@{}}Four LSTMs (C + R + W + Py)\\ Four BiLSTMs (C + R + W + Py)\\ RAFG\\ cw2vec(stroke-level)\end{tabular} & \begin{tabular}[c]{@{}c@{}}0.8072 (0.8078, 0.8074)\\ 0.8098 (0.8103, 0.8103)\\ 0.8181 (0.8181, 0.8187)\\ -- (--, --)\end{tabular} & \begin{tabular}[c]{@{}c@{}}0.7904 (0.7912, 0.7910) \\ 0.7915 (0.7925, 0.7921)\\ 0.7999 (0.7993, 0.8010)\\ -- (--, --)\end{tabular} & \begin{tabular}[c]{@{}c@{}}0.8826 (0.8841, 0.8811)\\ 0.8899 (0.8990, 0.8809)\\ 0.9172 (0.9201, 0.9144)\\ 0.9520 (0.9528, 0.9511)\end{tabular} & \begin{tabular}[c]{@{}c@{}}0.9018 (0.9022, 0.9014)\\ 0.9122 (0.9191, 0.9054)\\ 0.9002 (0.9033, 0.8972)\\ 0.9329 (0.9433, 0.9227)\end{tabular} \\ \hline
\begin{tabular}[c]{@{}c@{}}C-LSTMs (C)\\ C-LSTMs (C + R + W + Py)\\ C-BiLSTMs (C)\\ C-BiLSTMs (C + R + W + Py)\end{tabular} & \begin{tabular}[c]{@{}c@{}}0.8108 (0.8102, 0.8114)\\ 0.8163 (0.8177, 0.8149)\\ 0.8140 (0.8153, 0.8127)\\ 0.8211 (0.8246, 0.8177)\end{tabular} & \begin{tabular}[c]{@{}c@{}}0.7931 (0.7944, 0.7929)\\ 0.7956 (0.7951, 0.7972)\\ 0.7757 (0.7754, 0.7922)\\ 0.7939 (0.7957, 0.7922)\end{tabular} & \begin{tabular}[c]{@{}c@{}}0.8801 (0.8828, 0.8774)\\ 0.8823 (0.8775, 0.8871)\\ 0.9213 (0.9309, 0.9118)\\ 0.9264 (0.9384, 0.9147)\end{tabular} & \begin{tabular}[c]{@{}c@{}}0.9033 (0.9054, 0.9012)\\ 0.9036 (0.9068, 0.9004)\\ 0.9236 (0.9290, 0.9183)\\ 0.9294 (0.9332, 0.9257)\end{tabular} \\ \hline
Moto(BiLSTM)& \textbf{0.8316} (0.8346, 0.8287) & \textbf{0.8168} (0.8192, 0.8144) & \textbf{0.9638} (0.9671, 0.9605)&  \textbf{0.9633} (0.9679, 0.9588)\\ \hline
\end{tabular}
\label{tb:results}
\end{table*}

\begin{small}
\begin{figure*}[!ht]
    \centering
    \includegraphics[scale=0.55]{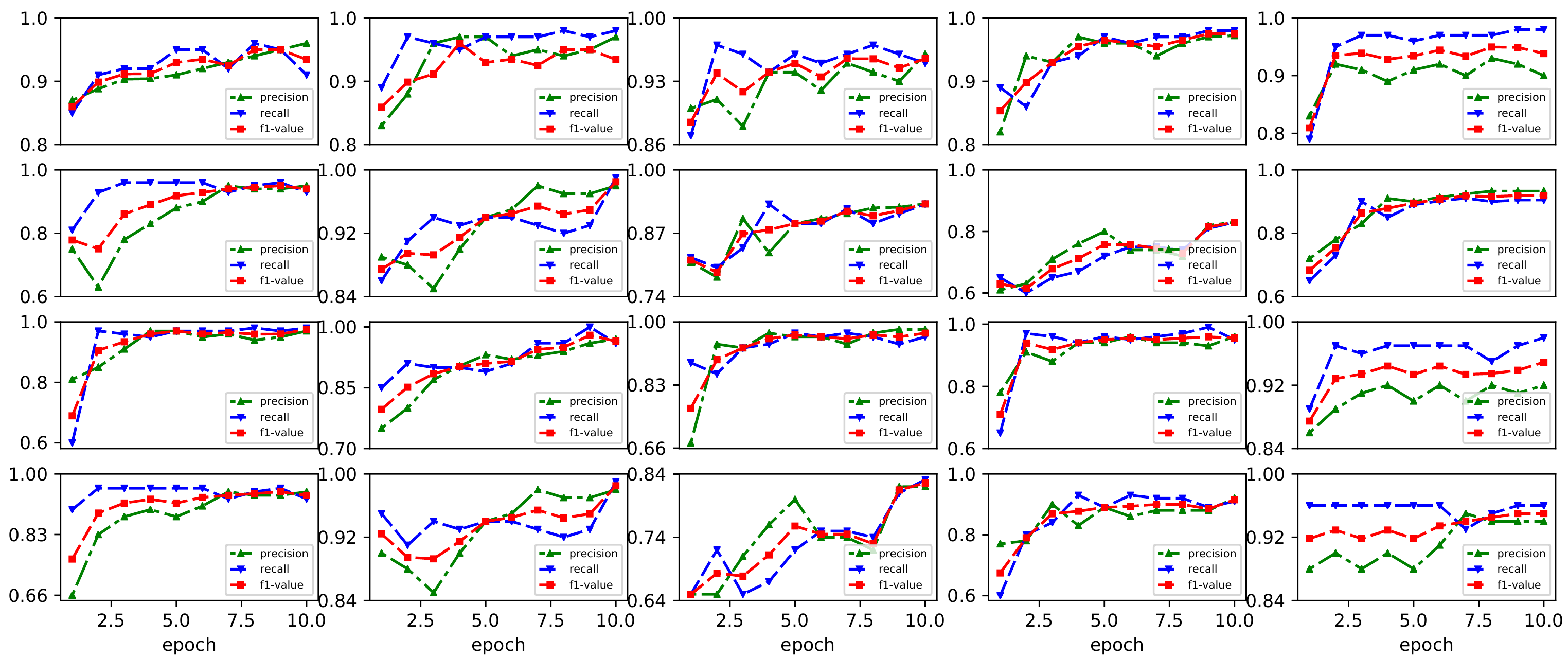}
    \caption{Details of the validations on the dataset of Fudan corpus, in which there are 20 classes, and xlabel refers to the number of epochs.}
    \label{fig:fudancorpus}
\end{figure*}
\end{small}

\subsection{Model Training} \label{sec:training}
Since what we are trying to solve is a multi-class classification task, we apply the cross-entropy \cite{zhou2016compositional,tao2019radical} loss function to train our model Moto, and the goal is to minimize the following {\it Loss}:
\vspace{-0.3cm}
\begin{small}
\begin{equation}
    \begin{split}
        Loss&= -\sum_{C\in Corpus}\sum_{i=1}^Kp_i(C)\log p_i(C) \\
    \end{split}
\end{equation}
\end{small}
where {\it C} is the input character-level input text. {\it Corpus} denotes the training corpus and {\it K} is the number of classes. In the period of training, we utilize {\it Adam} \cite{kingma2014adam} as optimizer to update the parameters of Moto. In addition, all {\it BiLSTMs} share properties, including weights and biases. 

\section{Experiments}
\subsection{Datasets}
To make a objectively comparison with baseline models, we conduct experiments separately on four datasets with gold standard classification labels. These four datasets include Chinese news titles (\#1 and \#2) \footnote{https://pan.baidu.com/s/1mgBTFOO}, Fudan corpus \footnote{https://github.com/yzwww2019/Fudan-corpus} , and THUCNews \footnote{http://thuctc.thunlp.org/message}.

\textbf{Dataset\#1} The dataset of Chinese news titles contains 47,952 titles with 32 classes ({\it i.e.,} the topics of news) for training and 15,986 titles for testing. In order to preserve the justice of the comporison with \cite{tao2019radical} and \cite{zhou2016compositional}, we do not filter out any texts.   

\textbf{Dataset\#2} To test the difference among these four aspects, we need to keep the purity of the dataset. To do so, we filter the original dataset{\it \#1} by removing the texts whose non-Chinese ratio is larger that 20\%, which refers to the approach proposed in \cite{tao2019radical}. The processed texts is remarked as dataset{\it \#2}.

\textbf{Dataset\#3} The dataset of Fudan corpus is a public dataset for Chinese text classification task. In this paper, we take 13,649 for training, and the remaining 4,549 for testing.

\textbf{Dataset\#4} The dataset of THUCNews contains 836,036 titles with 14 classes, 627,027 of which for training and the other 209,009 for testing. 

\subsection{Experimental Setup} \label{sec:es}
\textbf{Initial Setting}. We use {\it Open Chinese dictionary}, {\it Wubi library}, and {\it Pypinyin Library} to transform the character texts to radical, Wubi, and Pinyin texts, respectively, as discussed in Section \ref{sec: IL}. Moreover, we take the former average length {\it lavg} tokens into computing. And if the length of a sentence is smaller than {\it lvag}, we will use character `一' to make up the sentence until the length of it equals {\it lvag}.

\textbf{Embedding Setting}. Since the performance of deep learning models is highly related to the quality of input embeddings, we utilize the public word2vec tool ({\it Gensim}) to train embeddings for characters, radicals, Wubi codes, and Pinyin codes based on the large corpora \footnote{https://spaces.ac.cn/archives/4304}. The dimension of those embeddings are all set to 256 ({\it i.e., D=} 256). In addition, since the average length of texts in Fudan corpus and THUCNews is too longer for LSTMs to deal with, we take 1D-CNN \cite{kim2014convolutional} to convolute the texts in token dimension, ({\it i.e.,} the dimension of embedding is still kept at 256, but the token number is convoluted to ). For example, the original embedding matrix shape 32 $\times$ 4058 $\times$ 256 will be transformed to 32 $\times$ 18 $\times$ 256 as the input matrix shape. Finally, we conduct our experiment on 2 pieces of P100 GPU.

\textbf{Training Setting}. According to the previous training experience, we set the the dimension of hidden vectors in BiLSTM to 256, and set dropout rate to 50\% to escape overfitting. Moreover, the learning rate is set to 0.001 and we take Adam \cite{kingma2014adam} as optimizer for gradient descent computing. Furthermore, we set batch size to 32 empirically, and employ {\it Precision (P), Recall (R), and $F_1$-\textit{value}} to evaluate the performance \cite{hotho2005brief,qiao2019structure}, which is computed as follows:

\begin{small}
\begin{equation}
    F_1 = \frac{2\times(Precision\times Recall)}{Precision + Recall}
\end{equation}
\end{small}
\vspace{-0.3cm}
\subsection{Baseline Methods}
We compare our model with following baseline models in Chinese short text classification.

\begin{itemize}
  \item SVM+BOW. To evaluate the performance of radicals, Wubi codes, and Pinyin codes, we utilize tf-idf weights of characters ({\it C}), radicals ({\it R}), Wubi codes ({\it W}), and Pinyin codes ({\it Py}) as features separately, and train SVM classifier with {\it liblinear} \footnote{https://www.csie.ntu.edu.xn--tw/cjlin/liblinear/-784l}.
  \item Four LSTMs/BiLSTMs. We employ four LSTMs to process {\it C, R, W, and Py} as a whole baseline model, whose four corresponding hidden output would be concatenated into a vector. Similar to four LSTMs, we utilize four BiLSTMs as another baseline to test the effectiveness of bidirectional setting.
  \item RAFG \cite{tao2019radical}. RAFG is a four-granularity ({\it i.e.,} characters, radicals, character words, and radical words) model based on attention mechanism.
  \item cw2vec \cite{cao2018cw2vec}. cw2vec is a method for learning Chinese word embeddings in stroke-level information based on n-grams algorithm.
  \item C-LSTMs / C-BiLSTMs \cite{zhou2016compositional}. C-LSTMs employs two independent LSTMs to capture word and character features, which would be concatenated together. C-BiLSTMs is the bidirectional version of C-LSTMs.
\end{itemize}

\subsection{Experimental Results}
Table \ref{tb:results} demonstrates the {\it $F_1$-value}, Precision, and Recall of these baseline models and our Moto. In the following, we introduce these results in detail.

We provide the comparison results with SVM+BOW employing characters, radicals, Wubi codes, and Pinyin codes as features respectively. Table \ref{tb:results} shows that SVM + BOW (C) achieves the best average $F_1$-value 0.7955, 2.5\% higher than SVM + BOW (Py) in four Chinese text classification tasks. At the same time, Wubi gets average $F_1$-value 0.6954, as well radical gets 0.6554. The results indicate that all these four aspects are carriers of semantics in Chinese, and character plays the most important role in them.

\begin{small}
\begin{figure}[!ht]
    \centering
    \includegraphics[scale=0.33]{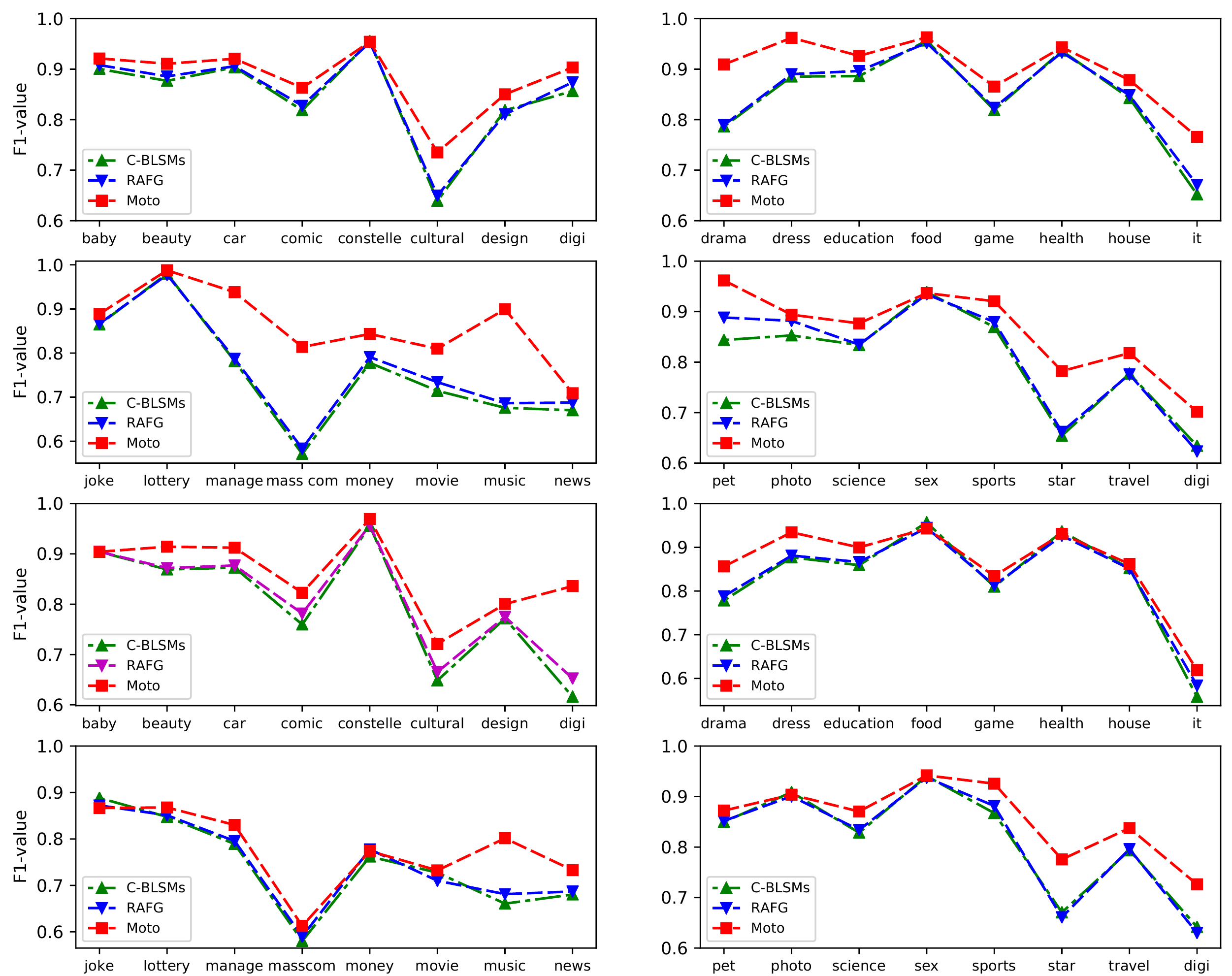}
    \caption{Detailed comparison on the dataset of Chinese news titles, Sub-figures in former two rows describe the dataset\#1, and sub-figures in the later two rows are related to dataset\#2.}
    \label{fig:Chinesenews}
\end{figure}
\end{small}

When comparing four LSTMs (C + R + W + Py), Four BiLSTMs (C + R + W + Py), RAFG, and cw2vec, we can find that RAFG which takes attention mechanism achieves the best performance, whose average $F_1$-value is 0.8589, higher than Four LSTMs (0.8455)) and Four BiLSTMs (0.8509). Moreover, cw2vec achieves the best performance in Fudan Corpus and THUCNews. Additionally, for C-LSTMs (C), C-LSTMs (C + R + W + Py), C-BiLSTMs(C), and C-BiLSTMs(C + R + W + P), the results indicate that methods with bidirectional version achieve better performance. At the same time, four-granularity model is better than single character-level model. Figure \ref{fig:Chinesenews} plots that the comparison in {\it $F_1$}-value among C-BiLSTMs, RAFG, and our model Moto. We can see that  Moto achieves the best performance in the most classes in dataset \#1 and dataset \#2.

For Moto, as shown in Figure \ref{fig:architecture}, we also employ four-granularity facts ({\it i.e,} but different from RAFG) based on attention mechanism. We conduct Moto on four datasets mentioned above. As a result, we can see that they achieve better score in {\it $F_1$}-\textit{value}, \textit{Precision}, and {\it Recall} compared with SVM+BOW, four LSTMs/BiLSTMs, RAFG, and cw2vec on datasets shown in Table \ref{tb:results}. In addition, Moto gets 0.9638 of $F_1$-value 1.24\% which is higher than the second method cw2vec in Fudan Corpus, and the detailed comparison in Fudan Corpus is shown in Figure \ref{fig:fudancorpus}. Finally, we conduct Moto in different aspact of these four granularities, and we can order the effectiveness of them as C $>$ Py $>$ R $>$ W.

\section{Conclusion}
We propose a novel method combining four granularities ({\it i.e., characters, radicals, Wubi codes, and Pinyin codes}) based on attention mechanism. Through the experiments on these four aspects, the order of importance in semantic of Chinese is that Moto (C) $>$ Moto (Py) $>$ Moto (R) $>$ Moto (W). In addition, the results in group Moto (C+X) ({\it X= R, W, or Py}) demonstrate that characters, radicals, Wubi codes, and Pinyin codes are unquestionably important semantic features in Chinese text classification. Our method Moto is 3.02\% higher the second method C-BiLSTM (C + R + W + Py) in average $F_1$-value in four datasets. Specifically, Moto improve the $F_1$-scores of four datasets: 1.28\%, 2.11\%, 1.24\%, and 3.26\%. In addition, our Moto achieves the SOTA in precision (1.89\% average improvement).

\bibliographystyle{IEEEtran}
\bibliography{IEEEabrv,IEEEexample}

\begin{thebibliography}{10}
\providecommand{\url}[1]{#1}
\csname url@samestyle\endcsname
\providecommand{\newblock}{\relax}
\providecommand{\bibinfo}[2]{#2}
\providecommand{\BIBentrySTDinterwordspacing}{\spaceskip=0pt\relax}
\providecommand{\BIBentryALTinterwordstretchfactor}{4}
\providecommand{\BIBentryALTinterwordspacing}{\spaceskip=\fontdimen2\font plus
\BIBentryALTinterwordstretchfactor\fontdimen3\font minus
  \fontdimen4\font\relax}
\providecommand{\BIBforeignlanguage}[2]{{%
\expandafter\ifx\csname l@#1\endcsname\relax
\typeout{** WARNING: IEEEtran.bst: No hyphenation pattern has been}%
\typeout{** loaded for the language `#1'. Using the pattern for}%
\typeout{** the default language instead.}%
\else
\language=\csname l@#1\endcsname
\fi
#2}}
\providecommand{\BIBdecl}{\relax}
\BIBdecl

\bibitem{shi2015radical}
X.~Shi, J.~Zhai, X.~Yang, Z.~Xie, and C.~Liu, ``Radical embedding: Delving
  deeper to chinese radicals,'' in \emph{Proceedings of ACL}, 2015, pp.
  594--598.

\bibitem{tao2019radical}
H.~Tao, S.~Tong, H.~Zhao, T.~Xu, B.~Jin, and Q.~Liu, ``A radical-aware
  attention-based model for chinese text classification,'' in \emph{AAAI},
  2019.

\bibitem{zhou2019multiple}
J.~Zhou, J.~Wang, and G.~Liu, ``Multiple character embeddings for chinese word
  segmentation,'' in \emph{Proceedings of ACL}, 2019, pp. 210--216.

\bibitem{chen2015neural}
S.~Chen, H.~Zhao, and R.~Wang, ``Neural network language model for chinese
  pinyin input method engine,'' in \emph{Proceedings of PACLIC}, 2015, pp.
  455--461.

\bibitem{peng2003text}
F.~Peng, X.~Huang, D.~Schuurmans, and S.~Wang, ``Text classification in asian
  languages without word segmentation,'' in \emph{Proceedings of IRAL}.\hskip
  1em plus 0.5em minus 0.4em\relax Association for Computational Linguistics,
  2003, pp. 41--48.

\bibitem{chen2015long}
X.~Chen, X.~Qiu, C.~Zhu, P.~Liu, and X.~Huang, ``Long short-term memory neural
  networks for chinese word segmentation,'' in \emph{Proceedings of EMNLP},
  2015, pp. 1197--1206.

\bibitem{cai2016neural}
D.~Cai and H.~Zhao, ``Neural word segmentation learning for chinese,'' in
  \emph{Proceedings of ACL}, 2016, pp. 409--420.

\bibitem{cai2017fast}
D.~Cai, H.~Zhao, Z.~Zhang, Y.~Xin, Y.~Wu, and F.~Huang, ``Fast and accurate
  neural word segmentation for chinese,'' in \emph{Proceedings of ACL}, 2017,
  pp. 608--615.

\bibitem{li2015component}
Y.~Li, W.~Li, F.~Sun, and S.~Li, ``Component-enhanced chinese character
  embeddings,'' in \emph{Proceedings of EMNLP}, 2015, pp. 829--834.

\bibitem{vaswani2017attention}
A.~Vaswani, N.~Shazeer, N.~Parmar, J.~Uszkoreit, L.~Jones, A.~N. Gomez,
  {\L}.~Kaiser, and I.~Polosukhin, ``Attention is all you need,'' in
  \emph{Advances in neural information processing systems}, 2017, pp.
  5998--6008.

\bibitem{mikolov2013distributed}
T.~Mikolov, I.~Sutskever, K.~Chen, G.~S. Corrado, and J.~Dean, ``Distributed
  representations of words and phrases and their compositionality,'' in
  \emph{Proceedings of NIPS}, 2013, pp. 3111--3119.

\bibitem{hochreiter1997long}
S.~Hochreiter and J.~Schmidhuber, ``Long short-term memory,'' \emph{Neural
  computation}, vol.~9, no.~8, pp. 1735--1780, 1997.

\bibitem{zhou2016compositional}
Y.~Zhou, B.~Xu, J.~Xu, L.~Yang, and C.~Li, ``Compositional recurrent neural
  networks for chinese short text classification.''\hskip 1em plus 0.5em minus
  0.4em\relax IEEE, 2016, pp. 137--144.

\bibitem{kingma2014adam}
D.~P. Kingma and J.~Ba, ``Adam: A method for stochastic optimization,''
  \emph{arXiv preprint arXiv:1412.6980}, 2014.

\bibitem{kim2014convolutional}
Y.~Kim, ``Convolutional neural networks for sentence classification,'' in
  \emph{Proceedings of EMNLP}, 2014, pp. 1746--1751.

\bibitem{hotho2005brief}
A.~Hotho, A.~N{\"u}rnberger, and G.~Paa{\ss}, ``A brief survey of text
  mining.'' in \emph{Ldv Forum}, vol.~20, no.~1.\hskip 1em plus 0.5em minus
  0.4em\relax Citeseer, 2005, pp. 19--62.

\bibitem{qiao2019structure}
L.~Qiao, H.~Zhao, X.~Huang, K.~Li, and E.~Chen, ``A structure-enriched neural
  network for network embedding,'' \emph{Expert Systems with Applications},
  vol. 117, pp. 300--311, 2019.

\bibitem{cao2018cw2vec}
S.~Cao, W.~Lu, J.~Zhou, and X.~Li, ``cw2vec: Learning chinese word embeddings
  with stroke n-gram information,'' in \emph{AAAI}, 2018.

\end{thebibliography}

\end{CJK}
\end{document}